\documentclass[runningheads]{llncs}
\usepackage{graphicx}
\usepackage{cite}
\usepackage{amsmath,amssymb,amsfonts}
\usepackage{algorithmic}
\usepackage{graphicx}
\usepackage{textcomp}
\usepackage{xcolor}
\usepackage{tikz}
\usepackage{subfigure}
\usepackage{arydshln}
\usepackage{pgfplots}
\usepackage{verbatim}

\usetikzlibrary{arrows, automata}
\usetikzlibrary{positioning}
\usetikzlibrary{arrows.meta}
\newcommand\copyrighttext{%
  D. Bacciu and A. Bruno, Deep Tree Transductions - A Short Survey \emph{To appear in the Proceedings of the 2019 INNS Big Data and Deep Learning (INNSBDDL 2019)}, Springer, 2019}
\newcommand\copyrightnotice{%
\begin{tikzpicture}[remember picture,overlay]
\node[anchor=north,yshift=-10pt] at (current page.north) {\fbox{\parbox{\dimexpr\textwidth-\fboxsep-\fboxrule\relax}{\copyrighttext}}};
\end{tikzpicture}%
}

\begin{document}
\title{Deep Tree Transductions - A Short Survey}
%
%
\author{Davide Bacciu\inst{1} \and
Antonio Bruno\inst{1}
}
\authorrunning{D. Bacciu and A. Bruno}
%
\institute{Department of Computer Science, University of Pisa, Pisa, Italy \email{\{bacciu,antonio.bruno\}@di.unipi.it}
}
\maketitle              
\copyrightnotice
\begin{abstract}
The paper surveys recent extensions of the Long-Short Term Memory networks to handle tree structures from the perspective of learning non-trivial forms of isomorph structured transductions. It provides a discussion of modern TreeLSTM models, showing the effect of the bias induced by the direction of tree processing. An empirical analysis is performed on real-world benchmarks, highlighting how there is no single model adequate to effectively approach all transduction problems.

\keywords{structured-data processing \and tree transduction \and TreeLSTM}
\end{abstract}
\section{Introduction}
Structured transductions are a natural generalization of supervised learning to application scenarios where both input samples and predictions are structured pieces of information. Trees are an example of non-trivial structured data which allows to straightforwardly represent compound information characterized by the presence of hierarchical-like relationships between its elements. Within this context, learning a tree structured transduction amounts to inferring a function associating an input tree to a prediction that is, as well, a tree. Many challenging real world applications can be addressed as tree transduction problems.

The problem of learning generic tree transductions, where both input and output trees have different topologies, is still a challenging open research question, despite some works \cite{tdLSTM}\cite{drnn} have started dealing with learning to sample tree-structured information, which is a prerequisite functionality for realizing a  predictor of tree structured outputs. Isomorphic transductions define a restricted form of tree transformations \cite{IOBUTHMM} which, nevertheless, allow to model and address several interesting learning tasks on structured data, including: (i) tree classification and regression, i.e. predicting a vectorial label for the whole tree; (ii) node relabeling, i.e. predicting a vectorial label for each node in a tree taking into account information from its surrounding context (e.g. its rooted subtree or its root-to-node path);
and (iii) reduction to substructure, i.e. predicting a tree obtained by pruning pieces of the input structure.

Isomorph transductions have been realized by several learning models, starting from the seminal work on the recursive processing of structures in \cite{generalframework}. There it has been formalized the idea of extending recurrent models to perform a bottom-up processing of the tree, unfolding the network recursively over the tree structure so that the hidden state of the current tree node depends from that of its children. The same approach has been taken by a number of models from different paradigms, such as the probabilistic bottom-up Hidden Markov Tree model \cite{buthmm}, the neural Tree Echo State Network \cite{treeesn} or the neural-probabilistic hybrid in Hidden Tree Markov Networks (HTNs) \cite{htn}. Another approach can be taken based on inverting the direction of parsing of the tree, by processing this top-down from the root to its leaves. This is diffused in particular in the probabilistic paradigm \cite{tdthmm}, where it represents a proper straightforward extension of hidden Markov models for sequences.  More recently, within the deep learning community it has diffused a widespread use of Long Short Term Memory (LSTM) networks \cite{lstm} for the processing of tree structured information. The so-called TreeLSTM model \cite{treelstmbu} extends the LSTM cell to handle tree-structures through a bottom-up approach implementing a specific instance of the recursive framework  by \cite{generalframework}. Although defined for the general case of an n-ary tree, the TreeLSTM has been used on binary trees obtained by binarization of the original parse tree which, in practice, reduces significantly the contribution of the structured information to solving the task. Two recent works show example of a top-down TreeLSTM:  one proposed in \cite{tdLSTM} to learn to sample trees and one in \cite{ssci18} showing the its use in learning simple non-isomorph transductions.

The aim of this paper is to present an orderly discussion of modern TreeLSTM models, assessing the effect of different stationarity assumptions (i.e. parameterization of the hidden state) on a full n-ary setting (i.e. without requiring a binarization of the structure). Also, since the choice of the tree parsing direction has a well-known effect on the representational capabilities of the model \cite{buthmm}, we consider both bottom-up and top-down TreeLSTM models. In particular, we focus on benchmarking the TreeLSTM models on three different tasks associated with the three different types of isomorph tree structured transductions discussed above. We will show how each of such task has different assumptions and characteristics which cannot all be effectively addressed by a single approach.

\section{Problem Formulation} \label{sect:back}
Before delving into the details of the different LSTM-based approaches to deal with tree data, we formalize the problems addressed in the experimental assessment using the generic framework of structured transductions.

We consider the problem of learning tree transductions from pairs of input-output trees $(\mathbf{x}^{n},\mathbf{y}^{n})$, where the superscript identifies the $n$-th sample pair in the dataset (omitted when the context is clear). We consider labeled rooted trees defined by the triplet ${(\mathcal{U}_{n},\mathcal{E}_{n},\mathcal{X}_{n})}$ consisting of a set of nodes $\mathcal{U}_{n} = \{1,\dots,U_n\}$, a set of edges $\mathcal{E}_{n}\subseteq\mathcal{U}_{n}\times\mathcal{U}_{n} $ and a set of labels $\mathcal{X}_{n}$. The term $u \in \mathcal{U}_{n}$ denotes a generic tree node whose direct ancestor, called \emph{parent}, is referred to as $pa(u)$. A node $u$ can have a variable number of direct descendants (\emph{children}), such that the $l$-th child of node $u$ is denoted as $ch_l(u)$. The pair $(u,v) \in \mathcal{E}_{n}$ is used to denote an edge between a generic node and its child and we assume trees to have maximum finite out-degree $L$ (i.e. the maximum number of children of a node). Each vertex $u$ in the tree is associated with a label $x_u$ ($y_u$, respectively) which can be of different nature depending on the application, e.g. a vector of continuous-valued features representing word embeddings or a symbol from a discrete and finite alphabet.

A tree transduction is defined as a mapping from an input sample to an output elements where both are tree structured pieces of information. Using $\mathcal{I}^{\#}, \mathcal{O}^{\#}$ to denote the input and output domains, respectively,  then a structural transduction is a function $\mathcal{F}: \mathcal{I}^{\#}\rightarrow \mathcal{O}^{\#}$. We focus on transductions exploiting the following definition of tree isomorphism.

\begin{definition}{Tree isomorphism}
Let $\mathbf{x} = {(\mathcal{U},\mathcal{E},\mathcal{X})}$ and $\mathbf{x}' = {(\mathcal{U'},\mathcal{E'},\mathcal{X'})}$, they are isomorphic if exists a bijection\\ $f:\mathcal{U} \rightarrow \mathcal{U'}$ such that $ \forall (u,u')\in \mathcal{E} \iff (f(u), f(u'))\in \mathcal{E'}$.
\end{definition}

An equivalent definition can be given using the concept of \emph{skeleton}.
\begin{definition}{Skeleton tree}
Let $\mathbf{x} = {(\mathcal{U},\mathcal{E},\mathcal{X})}$, its skeleton is  $skel(\mathbf{x}) = {(\mathcal{U},\mathcal{E})}$.
\end{definition}
Following such definition, two trees are isomorphic if they have the same skeleton (labels are irrelevant, only structure matters).

A general \emph{structured transduction} can be formalized by a learnable encoding-decoding process where $\mathcal{F} = \mathcal{F}_{out} \circ \mathcal{F}_{enc}$ with:
\begin{equation*}
\mathcal{F}_{enc}: \mathcal{I}^{\#}\rightarrow \mathcal{H}^{\#} \qquad \qquad \mathcal{F}_{out}: \mathcal{H}^{\#}\rightarrow \mathcal{O}^{\#}.
\end{equation*}
The terms $\mathcal{F}_{enc}$ and $\mathcal{F}_{out}$ are the \emph{encoding} and \emph{output} transductions, while we assume the existence of a state space $\mathcal{H}^{\#}$ providing an intermediate and rich representation of the structured information, such as in the activations of the hidden neurons of a recursive neural model. Different types of transductions can be obtained depending on the isomorphism properties of the encoding and output mappings. In this work, we consider three types of tree transductions, each associated with a practical learning and prediction task. First, we consider a \emph{tree-to-tree isomorphic transduction} where both encoding and output mappings are isomorphic. The second type is the \emph{structure-to-element} or \emph{supersource transduction} that map an input tree into a single vectorial element in the output domain, basically realizing a classical tree classification or regression task. The third type is the  \emph{structure-to-substructure} transduction which defines a restricted form of non-isomorphic transduction where the output tree $\mathbf{y}$ is obtained from the input tree $\mathbf{x}$ by pruning some of its proper subtrees. In practice, such a transformation can again be realized as an isomorphic transduction, where the  encoding is isomorphic as in the previous cases. The output function, instead, isomorphically maps each element of the hidden encoding into an output node while  using a specific \emph{NULL} value as label of those nodes of the input structure that are non-existing in the output tree.

\section{TreeLSTM for Constrained Tree Transductions} \label{sect:mod}
Several works have been dealing with extending Recurrent Neural Networks (RNN) to deal with tree structured data. Lately, most of these works focused on tree-structured extensions of LSTM cells and networks. Two sources of differentiation exist between the different models. One concerns the stationarity assumptions, that is how much tied are the network parameters with respect to topological aspects such as the position of a node with respect to its siblings. The second source of differentiation concerns the direction of tree processing (top-down or bottom-up for trees), which determines the context upon which a specific node is assessed (i.e. depending on the hidden state of the parent, for the top-down case, or depending on the states of its children, for the bottom-up case). In the following, we briefly review the TreeLSTM approaches in literature with respect to these two differentiating factors.

\begin{description}
\item[Top-Down (TD) TreeLSTM:]in this model, tree processing flows from the root to the leaves. In literature, TD TreeLSTM models are mainly used in generative settings, where one wants to generate the children of a node based on the hidden state of the parent \cite{tdLSTM,drnn}. Their use as encoders of the full structure is not common, as this requires some form of mapping function summarizing the whole tree into a single encoding vector (e.g. the mean of the hidden states of all the nodes in the tree).
Here, we consider the use of TD TreeLSTM in the context of learning isomorph transductions of the three types discussed in Section \ref{sect:back}. In particular, we will assess the capabilities of this model in realizing non-generative tasks, highlighting limitations and advantages with respect to its more popular bottom-up counterpart. Formally, the activation of a TD TreeLSTM cell for a generic node $u$ is regulated by the following equations
\begin{align}
 r_u &= \tanh \biggl( W^{(r)} x_u + U^{(r)} h_{pa(u)} + b^{(r)} \biggr) 
\\
 i_u &= \sigma \biggl( W^{(i)} x_u + U^{(i)} h_{pa(u)} + b^{(i)} \biggr) 
\\
 o_u &= \sigma \biggl( W^{(o)} x_u + U^{(o)} h_{pa(u)} + b^{(o)} \biggr) 
\\
 f_u &= \sigma \biggl( W^{(f) }x_u + U^{(f)} h_{pa(u)} + b^{(f)} \biggr) 
\\\
 c_u &= i_u \odot r_u+ f_u \odot c_{pa(u)} 
\\
 h_u &= o_u \odot \tanh (c_u) 
\end{align}
\noindent with the term $x_u$ denoting unit input, $h_{pa(u)}$ and $c_{pa(u)}$ are, respectively, the hidden state and the memory cell state of the node's parent, $\sigma$ is the sigmoid activation function and $\odot$ is elementwise multiplication. It can be seen that the formal model of this unit is that of a standard LSTM unit for sequences, but this will be unfolded over the tree in a TD fashion by following in parallel all the root to leaves paths.
\end{description}
The second type is Bottom-Up (BU) TreeLSTM, in which tree processing flows from the leaves to the root. In literature there are two types of BU TreeLSTM, which is mainly used as one-pass encoder for tree structured information \cite{treelstmbu}. The two BU TreeLSTM types differ in stationarity assumptions and the choice of which one to use depends on the specificity of the structured data at hand (e.g. finiteness of the outdegree,  relevance of node positionality information). The choice of a bottom-up approach is motivated by consolidated results showing a superior expressiveness of bottom-up parsing with respect top-down approaches when dealing with trees \cite{buthmm}. When considered within the context neural processing of the structure this founds on the assumption that a node hidden state computed recursively from its children states is a ``good" vectorial summary of the information in all the subtree rooted in the node. Another observation is that the bottom-up approach provides a natural means to obtain a state mapping function for the whole tree, by considering the hidden state of the tree as a good summary of the information contained in the whole structure.
\begin{description}
\item[Child-Sum TreeLSTM:] this type of TreeLSTM is used to encode trees where the position of nodes (ordering) with respect to their siblings is not relevant for the task. Let $ch(u)$ be the set of children (of size $K$) of the generic node $u$, its state transition equation follows:
\begin{align} 
\tilde h_u &= \sum_{k \in ch(u)} h_k
\\
 r_u &= \tanh \biggl( W^{(r)} x_u + U^{(r)} \tilde h_u + b^{(r)} \biggr)
\\
 i_u &= \sigma \biggl( W^{(i)} x_u + U^{(i)} \tilde h_u + b^{(i)} \biggr)
\\
 o_u &= \sigma \biggl( W^{(o)} x_u + U^{(o)} \tilde h_u + b^{(o)} \biggr)
\\
 f_{uk} &= \sigma \biggl( W^{(f) }x_u + U^{(f)} h_k + b^{(f)} \biggr) , \forall k \in ch(u)
\\
 c_u &= i_u \odot r_u +  \sum_{k \in ch(u)} f_{uk} \odot c_k
\\
 h_u &= o_u \odot \tanh (c_u)
\end{align}

\noindent with the term $x_u$ denoting unit input, $h_k$ and $c_k$ are, respectively, the hidden state and the memory cell state of the $k$-th child, $\sigma$ sigmoid function and $\odot$ the elmentwise product. As every LSTM it has the three gates, in particular there is a forget gate for every child but all of them share the same parameters.\\

\item[N-ary TreeLSTM:] this TreeLSTM variant allows to discriminate children by their position with respect to the siblings, while needing to fix a priori the maximum outdegree of the tree. Let $u$ be the generic node, the associated TreeLSTM cell activations are as follows
\begin{align} 
 r_u &= \tanh \biggl( W^{(r)} x_u +  \sum_{\ell=1}^N U_\ell^{(r)} h_{ch_{\ell}(u)} + b^{(r)} \biggr)
\\
 i_u &= \sigma \biggl( W^{(i)} x_u +  \sum_{\ell=1}^N U_\ell^{(i)} h_{ch_{\ell}(u)} + b^{(i)} \biggr)
\\
 o_u &= \sigma \biggl( W^{(o)} x_u +  \sum_{\ell=1}^N U_\ell^{(o)} h_{ch_{\ell}(u)} + b^{(o)} \biggr)
\\
\begin{split}
f_{uk} &= \sigma \biggl( W^{(f)} x_u +  \sum_{\ell=1}^N U_{k\ell}^{(f)} h_{ch_{\ell}(u)} + b^{(f)} \biggr), \\ \forall k &= 1,2,\ldots,N  \label{forgetgatepar}
\end{split}
\\
 c_u &= i_u \odot r_u +  \sum_{\ell=1}^N f_{u\ell} \odot c_{ch_{\ell}(u)}
\\
 h_u &= o_u \odot \tanh (c_u)
\end{align}
\noindent with the term $x_u$ denoting unit input, $h_{ch_{\ell}(u)}$ and $c_{ch_{\ell}(u)}$ are, respectively, the hidden state and the memory cell state of the $\ell$-th child, $\sigma$ is the sigmoid function and $\odot$ is the element-wise product.

The introduction of separate parameter matrices for each child allows the model to learn more fine-grained conditioning on the states of a unit's children. Equation \eqref{forgetgatepar} shows a parametrization of the $k$-th child's forget gate $f_{uk}$ that allows more flexible control of information propagation from child to parent and it can be also used to control the influence between siblings.
\end{description}

\section{Experiments and Results} \label{sect:exp}
In this section, we empirically evaluate the TreeLSTM types surveyed in Section \ref{sect:mod} in three different classes of tree transduction tasks. All tasks are based on real-world data and have been chosen as they allow to further compare with other approaches in literature. Model selection choices have been performed on hold-out validation data and final performance is assessed on further hold-out test data. A $L_{2}$ penalization term has been added to the loss function for the sake of model regularization, using a penalization weight fixed to $\lambda = 10^{-4}$. The only hyperparameter in model selection is the number of LSTM units, chosen in $\{ 100, 150, 200, 250, 300, 350, 400\}$, and we have used a standard Adam optimizer \cite{adam}. For the sake of compactness, we only report test-set results, obtained by averaging on 10 independent runs with random weights initializations (variance is not reported, if less than millesimal).

The first experiment assesses TreeLSTM models on structure-to-element transductions by means of two tree classification tasks coming from the INEX 2005 and INEX 2006 competitions \cite{Inex} (task INEX20xx in the following). Both dataset are multiclass classification tasks based on trees that represent XML documents from different thematic classes. The datasets are provided with standard splits in training and test sets: for model selection purposes we held out 10\% of the former to define the validation set (by stratification, to preserve the dataset class ratios in the folds). All TreeLSTM models used a LogSoftMax layer in output and a Negative Log-Likelihood loss. Performance on both INEX20xx datasets are evaluated in terms of classification accuracy, reported in Table \ref{tab:inex05} and \ref{tab:inex06} for the model-selected configurations of each TreeLSTM model. These results highlight how the BU approaches outperform TD approaches in structure-to-element transductions. This is due to the fact that a BU encoding results in node hidden activations that summarize information concerning the whole subtree rooted on the node. When compared to other state-of-the-art approaches, the TreeLSTMs show very competitive results in INEX 2005 (where the best accuracy is $97.15\%$, attained by PAK-PT \cite{bestinex05}) while the ChildSum TreeLSTM has the best performance in INEX 2006 (the runner up being the Jaccard kernel in \cite{bestkernel06}, with an accuracy of $45.06\%$).
\begin{table}[tb]
\begin{minipage}{0.46\textwidth}
\caption{Inex05 Test Results}
\centering
\begin{tabular}{c}
\newcommand\T{\rule{0pt}{2.5ex}}       
\resizebox{\textwidth}{!}
{
\begin{tabular}{l|c}
\multicolumn{1}{c|}{\bf{Model}} & \bf{Accuracy \%}\\
\hline \T
TD-TreeLSTM & 62.05\\
ChildSum TreeLSTM & 82.85\\
\bf{N-ary TreeLSTM} & \bf{96.89} \\
\end{tabular}
}
\label{tab:inex05}
\end{tabular}

\end{minipage}
\begin{minipage}{0.49\textwidth}
\caption{Inex06 Test Results}
\centering
\begin{tabular}{c}
\newcommand\T{\rule{0pt}{2.5ex}}       
\resizebox{\textwidth}{!}
{
\begin{tabular}{l|c}
\multicolumn{1}{c|}{\bf{Model}} & \bf{Accuracy \%}\\
\hline \T
TD-TreeLSTM & 25.07\\
\bf{ChildSum TreeLSTM} & \bf{46.12}\\
N-ary TreeLSTM & 38.57 \\
\end{tabular}
}
\label{tab:inex06}
\end{tabular}
\end{minipage}
\end{table}

The second experiment is focused on the assessment of TreeLSTM models on structure-to-substructure transductions based on the CLwritten corpus \cite{CLcorpora}. This is a benchmark dataset for sentence compression techniques which is based on sentences from written sources whose ground truth shortened versions  were created manually. Here, the annotators were asked to produce the smallest possible target compression by deleting extraneous words from the source, without changing the word order and the meaning of the sentences.

The original corpus provides sentences in sequential form (see \cite{CLcorpora} for more details). The corresponding tree representation has been obtained using the (constituency) \emph{Stanford Parser} \cite{stanfordparser} on the original sentences. Node labels in the resulting trees are of two kind: leaves are labeled with vocabulary words, represented through \emph{word embeddings} obtained using the \emph{word2vec} \cite{word2vec}. Internal nodes are labelled with semantic categories that are, instead, represented using a one-hot encoding. For model selection and validation purposes we have split the corpus in $903$ training trees, $63$ validation samples and $882$ test trees, along the lines of the experimental setup defined for the baseline models in \cite{LapataCompress,compressLSTM}.

The TreeLSTM output is generated by a layer of sigmoid neurons, where an activation smaller than $0.5$ represents the fact that the corresponding input node is not present in the output tree, while the opposite means that the node (and the corresponding word) is preserved. The associated loss is, indeed, the Binary Cross Entropy. Performance on the corpus has been assessed using two metrics assessing different aspects of compression quality:
\begin{description}
\item[accuracy or importance factor:] this metric measures how much of the important information is retained. Accuracy is evaluated using \emph{Simple String Accuracy} (SSA) \cite{ssa}, which is based on the string edit distance between the compressed output generated by the model and the reference ground-truth compression;
\item[compression rate:] this metric measures how much of the original sentence remains after compression. Compression rate is defined as the length of the compressed sentence divided by the original length, so lower values indicate more compression;
\end{description}
During training, early stopping decisions are taken by considering an hybrid metric, which trades off accuracy and compression according to the following definition
\[
t = \frac{\text{accuracy}^2}{\text{compression rate}}.
\]

Table \ref{tab:compression} reports the performance values for the CLWritten task. Here it is evident that the TD approach outperforms both BU approaches. As in the first experiment, this is due to the characteristics of the transduction task. In a parse tree the words occur only at leaves and a disambiguation of their interpretation can be performed only by considering their context, which can only come from their ancestors because they have no children by definition. Hence, it follows that a parent-to-children information flow is more relevant than a children-to-parent one to determine if a word, represented necessarily by a leaf node, has to be included or not in a summary. The best result in literature for the task is obtained by a LSTM applied to the original sequential representation of the data, achieving less than $70\%$ in accuracy and about $82\%$ in compression. This marks the clear advantage of using a TD tree transduction approach, as the corresponding TreeLSTM outperforms the sequential model both in accuracy and compression.
\begin{table}[tb]
\begin{minipage}{\textwidth}
\caption{CLWritten summarization results: the reference compression value for the gold-standard compressions is $70.41$.}
\centering
\begin{tabular}{c}
\newcommand\T{\rule{0pt}{2.5ex}}       
\begin{tabular}{l|c|c}
\multicolumn{1}{c|}{\bf{Model}} & \bf{Accuracy \%} & \bf{Compression \%}\\
\hline \T
\bf{TD-TreeLSTM} & \bf{73.58} & \bf{72.37}\\
ChildSum TreeLSTM & 63.17 & 84.01\\
N-ary TreeLSTM & 63.41 & 91.46\\
\end{tabular}
\label{tab:compression}
\end{tabular}
\vspace{.4cm}
\end{minipage}
\end{table}

The last experiment assesses the TreeLSTM models on an isomorph structure-to-structure transduction. The task requires to relabel an input tree into an isomorphic structure with changed labels. This is applied to the problem of inferring the semantic categories of a parse tree (grammar induction), given the structure of the parse tree and knowledge on the labels of the leaves (i.e. the words in the sentences). To this end, we have used the Treebank2 dataset focusing, in particular, on the Wall Street Journal (WSJ) subset of trees. Due to the fact that output labels can be assigned taking values from a discrete set, the output layers used to label internal nodes is of LogSoftMax type (with Negative Log-Likelihood as training loss). For our experiments, we used the standard WSJ partition, where sections 2-21 are used for training, section 22 for validation, and section 23 for testing. The performance metric for the task is label accuracy (computed as the proportion of correctly inferred semantic categories for each parse tree). Table \ref{tab:treebank} reports the resulting accuracies for the TreeLSTM models:  we were not able to obtain results for the N-ary TreeLSTM configuration within reasonable computing time ($2$ weeks) for this task, due to its computational complexity. Nevertheless, results highlight how the BU approach based on the ChildSum encoding outperforms the TD one. As in the first experiment, relevant information on this task flows from the leaves to the root, i.e. from the sentence words, which are observable, to the internal nodes, where we want to infer the missing semantic categories.
\begin{table}
\begin{minipage}{\textwidth}
\caption{Treebank2 Test Results}
\centering
\begin{tabular}{c}
\newcommand\T{\rule{0pt}{2.5ex}}       
\begin{tabular}{l|c}
\multicolumn{1}{c|}{\bf{Model}} & \bf{Test Accuracy \%}\\
\hline \T
TD-TreeLSTM & 49.56\\
\bf{ChildSum TreeLSTM} & \bf{95.23}\\
\end{tabular}
\label{tab:treebank}
\end{tabular}
\end{minipage}
\end{table}

Comparison with the start-of-the-art can't be done due to the fact we used a restricted set of the output labels, so comparisons would be unfair.

\section{Concluding Remarks} \label{sect:conc}
We surveyed different TreeLSTM architectures and parameterizations, providing an empirical assessment of their performance on learning different types of constrained tree transductions. Our analysis, not unsurprisingly, concluded that there is no single best configuration for all transduction problems. In general, one must choose the model which computes the structural encoding following a direction of elaboration akin to the information flow in the structure. This said, in the majority of cases, the BU approach proved more effective than the TD, thanks to the fact that in BU approaches, the state of a node summarizes the information of its rooted subtree. Moreover, we showed that TreeLSTMs have competitive performances with respect to the state-of-the-art models in all types of structured transductions.

\section*{Acknowledgment}
This work has been supported by the Italian Ministry of Education, University, and Research (MIUR) under project SIR 2014 LIST-IT (grant n. RBSI14STDE).

%
%

\end{document}